\documentclass[11pt,a4paper]{article}
\usepackage[hyperref]{emnlp-ijcnlp-2019}
\usepackage{times}
\usepackage{latexsym}

\usepackage{color}
\usepackage{booktabs}
\usepackage{tabularx}
\usepackage{graphicx}
\usepackage{subcaption}
\usepackage{todonotes}

\usepackage{url}

\usepackage[utf8]{inputenc}
\usepackage[T1]{fontenc}

\aclfinalcopy 

\newcommand{\affut}{$^1$}
\newcommand{\afftilde}{$^2$}

\title{Grammatical Error Correction and Style Transfer via\\Zero-shot Monolingual 
Translation}

\author{
    Elizaveta Korotkova\affut,
    Agnes Luhtaru\affut,
    Krista Liin\affut,
    Maksym Del\affut,
    Daiga Deksne\afftilde,
    Mark Fishel\affut \\~\\
  \affut Institute of Computer Science, University of Tartu, Estonia \\
  \texttt{name.surname@ut.ee} \\~\\
  \afftilde Tilde, Riga, Latvia\\
  \texttt{daiga@deksne.lv}}

\newcommand{\sctref}[1]{Section~\ref{#1}}
  
\date{}

\begin{document}
\maketitle

\begin{abstract}
Both grammatical error correction and text style transfer can be viewed as monolingual sequence-to-sequence transformation tasks, but the scarcity of directly annotated data for either task makes them unfeasible for most languages. We present an approach that does both tasks for multiple languages using the same trained model, while only using regular language parallel data and without requiring error-corrected or style-adapted texts. We apply our model to three languages and present a thorough automatic and manual evaluation on both tasks, showing that the proposed approach is reliable for a number of error types and style transfer aspects.
\end{abstract}

\section{Introduction}







Sequence-to-sequence (seq2seq) transformations have recently proven to be a successful framework
for several natural language processing tasks, like machine translation (MT)
\cite{bahdanau,transformer}, speech recognition
\cite{asr}, speech synthesis \cite{tts}, natural language inference \cite{parikh-emnlp} 
and others. However, the success of these models depends on the availability of large amounts of directly annotated data for the task at hand (like translation examples, text segments and their speech recordings, etc.). This is a severe limitation for tasks where data is not abundantly available as well as for low-resource languages.

Here we focus on two such tasks: grammatical error correction (GEC) and style transfer. Modern approaches to GEC learn from parallel corpora of erroneous segments and their manual corrections \cite{conll2014,gecnmt}; text style transfer also relies on supervised approaches that require texts of the same meaning and different styles \cite{sup1,sup2} or imprecise unsupervised methods \cite{g1,umt1}.

In this paper we introduce an approach to performing both GEC and style transfer with the same trained model for multiple languages, while not using any supervised training data for either task. It is based on zero-shot neural machine translation (NMT) \cite{google-zero-shot}, and as such, the only kind of data it uses is regular parallel corpora (with texts and their translations). However, we apply the model to do \emph{monolingual} transfer, translating the input segment into the same language; we show, that this ``monolingual translation'' is what enables the model to correct input errors and adapt the output into a desired style.

Our main contributions are thus: \textit{(i)} a method for both style transfer and grammatical error correction that does not use annotated training data for either task, \textit{(ii)} support for both tasks on multiple languages within the same model,
\textit{(iii)} a thorough quantitative and qualitative manual evaluation of the model on both tasks, and \textit{(iv)} highlighting of the model's reliability aspects on both tasks.
We used publicly available software and corpora in this work; an online demo of our results is available, but concealed for anonymization purposes.

We describe the details of our approach in \sctref{method}, then evaluate it in terms of style transfer performance in \sctref{style} and grammatical error correction performance \sctref{gec}. The paper ends with a review of related work in \sctref{relwork} and conclusions in \sctref{conclusions}.

\section{Method}
\label{method}

\begin{figure}
\begin{center}
\includegraphics[width=170pt]{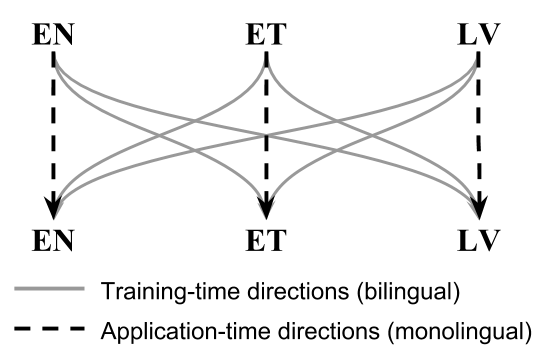}
\end{center}
\caption{Schematic illustration of zero-shot monolingual translation. The model is trained on bilingual data in all translation directions (English-to-Estonian, Estonian-to-English, English-to-Latvian, etc.) and then applied in monolingual directions only (English-to-English, etc.), without having seen any sentence pairs for them. The illustration is simplified, as it does not show the style (text domain) parametrization.}
\label{figMMNMT}
\end{figure}

Our approach is based on the idea of zero-shot MT \cite{google-zero-shot}. There the authors show that after training a single model to translate from Portuguese to English as well as from English to Spanish, it can also translate Portuguese into Spanish, without seeing any translation examples for this language pair. We use the zero-shot effect to achieve \emph{monolingual translation} by training the model on bilingual examples in both directions, and then doing translation into the same language as the input: illustrated on Figure~\ref{figMMNMT}.

With regular sentences monolingual translation does not seem useful, as its behaviour mainly consists of copying. However, when the input segment has characteristics rarely or never seen by the model at training time (like grammatical errors or different stylistic choices) -- the decoder still generates the more regular version of the sentence (thus fixing the errors or adapting the style). Furthermore, in case of multilingual multi-domain NMT \cite{tars}, it is possible to switch between different domains at runtime, thus doing style transfer via monolingual adaptation.

To create a multilingual multi-domain NMT system we use the self-attention architecture \cite[``Transformer'',][]{transformer}. Instead of specifying the output language with a token inside the input sequence, as \citet{google-zero-shot} did, we follow \cite{tars} and use word features (or \emph{factors}). On one hand, this provides a stronger signal for the model, and on the other -- allows for additional parametrization, which in our case is the text domain/style of the corpus.
As a result, a pre-processed English-Latvian training set sentence pair \emph{``Hello!''--``Sveiki!''} looks like:\\
\begin{tabular}{ll}
& \\
\textbf{En:}& \verb%hello|2lv|2os !|2lv|2os%\\
\textbf{Lv:}& \verb%sveiki !%\\
& \\
\end{tabular}\\
\noindent Here \verb|2lv| and \verb|2os| specify Latvian and OpenSubtitles as the output language and domain; the output text has no factors to predict. At application time we simply use the same input and output languages, for example the ungrammatical input \emph{``we is''} looks like the following, after pre-processing:\\
\begin{tabular}{ll}
& \\
\textbf{En:} & \verb%we|2en|2os is|2en|2os%\\
& \\
\end{tabular}

The intuition behind our approach is that a multilingual shared encoder produces semantically rich latent sentence representations \cite{laser}, which provide a solid  ground for the effective style transfer on top.

Next we present the technical details, the experiment setup and the data we used for training the model used in the experiments.

\begin{table*}[t]
\centering
\captionof{table}{Examples of style transfer}
\begin{tabular}{c|cc}
\toprule
\multicolumn{2}{c}{Translation into informal style (OpenSubtitles)}\\
\midrule
I could not allow him to do that. & I couldn't let him do that. \\
He will speak with Mr. Johns. & He'll talk to Mr. Johns.\\
I will put you under arrest. & I'll arrest you.\\
\toprule
\multicolumn{2}{c}{Translation into formal style (Europarl)}\\
\midrule
How come you think you're so dumb? & Why do you think you are so stupid?\\
I've been trying to call. & I have tried to call.\\
Yeah, like I said. & Yes, as I said.\\
\bottomrule
\end{tabular}
      \label{tab-examples}
\end{table*}

\subsection{Languages and Data}

We use three languages in our main experiments: English, Estonian and Latvian\footnote{Preliminary results with bilingual training with just two languages fed in both directions does not lead to reliable monolingual translation, since despite requesting translations into the same language as the input, the model learns to ``guess'' the output language based on the input language and often ignores the output language factor, instead performing cross-lingual translation regardless of its value.}. All three have different characteristics: Latvian and (especially) Estonian are morphologically complex and have loose word order, while English has a strict word order and the morphology is much simpler. Most importantly, all three languages have error-corrected corpora for testing purposes, though work on GEC for Estonian and Latvian is extremely limited (see \sctref{gec}).

The corpora we use for training the model are OpenSubtitles2018 \cite{opensubs}, Europarl \cite{europarl}, JRC-Acquis and EMEA \cite{europarl-jrc}. We assume that there should be sufficient stylistic  difference  between  these  corpora,  especially  between  the  more  informal  OpenSubtitles2018 (comprised of movie and TV subtitles)  on one hand and Europarl and JRC-Acquis (European parliamentary speeches and legal texts) on the other.\footnote{We  acknowledge the fact that most text corpora and OpenSubtitles in particular constitute a heterogeneous mix of genres and text characteristics; however, many stylistic traits are also similar across the whole corpus, which means that these common traits can be learned as a single style.}

\subsection{Technical Details}

For Europarl, JRC-Acquis and EMEA we use all data available for English-Estonian, English-Latvian and Estonian-Latvian language pairs. From OpenSubtitles2018 we take a random subset of 3M sentence pairs for English-Estonian, which is still more than English-Latvian and Estonian-Latvian (below 1M; there we use the whole corpus). This is done to balance the corpora representation and to limit the size of training data.

Details on the model hyper-parameters, data pre-processing and training can be found in Appendix~\ref{apx1}.

\subsection{Evaluation}

First, we evaluate our model in the context of MT, as the translation quality can be expected to have influence on the other tasks that the model performs. We use public benchmarks for Estonian-English and Latvian-English translations from the news translation shared tasks of WMT 2017 and 2018 \cite{wmt17,wmt18}. The BLEU scores for each translation direction and all included styles/domains are shown in Table~\ref{tab-bleu}.
\begin{table}[h]
\centering
\captionof{table}{BLEU scores of the multilingual MT model on WMT'17 (Latvian$\leftrightarrow$English) and WMT'18 (Estonian$\leftrightarrow$English) test sets}
\begin{tabular}{c|cccc}
\toprule
      & to EP & to JRC  & to OS & to EMEA\\
\midrule
EN$\rightarrow$ET & \textbf{20.7} & 19.9 & 20.6 & 18.6\\
ET$\rightarrow$EN & 24.7 & 23.6 & \textbf{26.1} & 23.8\\
EN$\rightarrow$LV & 15.7 & 15.3 & \textbf{16.3} & 15.0\\ 
LV$\rightarrow$EN & 18.3 & 17.8 & \textbf{19.0} & 17.5\\
\bottomrule
\end{tabular}
      \label{tab-bleu}
\end{table}

Some surface notes on these results: the BLEU scores for translation from and into Latvian are below English-Estonian scores, which is likely explained by smaller datasets that include Latvian. Translation into English has higher scores than into Estonian/Latvian, which is also expected.

An interesting side-effect we have observed is the model's ability to handle code-switching in the input. The reason is that the model receives only the target language (and domain) as additional input, and not the source language, and as a result it learns language normalization of sorts. For example, the sentence ``Ma tahan two saldējumus.'' (\emph{``Ma tahan''} / \emph{``I want''} in Estonian, \emph{``two''} and \emph{``saldējumus''} / \emph{``ice-creams''} in genitive, plural in Latvian) is correctly translated into English as ``I want two ice creams.''. See more examples in Appendix~\ref{apx2}. The same sentence, when translated into Estonian or Latvian results in the grammatically correct \emph{``Ma tahan kahte jäätist.''} / \emph{``Es gribu divus saldējumus.''}.

\section{Style Transfer}
\label{style}

In this section we evaluate our model's performance in the context of style transfer.
The assumption is that passing modified style factors should prevent the model from simply copying the source sequences when translating inside a single language, and incentivize it to match its output to style characteristics typical for different corpora.
To assess whether that is the case, 
we performed automatic and manual evaluation.

We limit further comparisons to two styles, translating sentences of the OpenSubtitles test set into the style of Europarl and vice versa, designating them as ``informal'' and ``formal'' texts for the purpose of this work. We assume that, generally, movie subtitles gravitate towards the more informal style, and parliament proceedings towards the more formal (see examples of translations into those styles in Table~\ref{tab-examples}). Preliminary tests showed that JRC-Acquis and EMEA texts resulted in practically the same style as Europarl. We also leave Estonian and Latvian out of the evaluations, as there are neither corpora nor prior style transfer solutions in these languages for performing  comparisons; some output examples for these languages are given in Appendix~\ref{apx2}.

\subsection{Automatic Evaluation}
\label{auto-eval}

To evaluate the strength of style transfer quantitatively, we compare our system to that of \citet{sup5}. They train MT systems on parallel informal texts and their formal rewrites, which were manually created specifically for their collected corpus (titled GYAFC). This approach, while relying on style-parallel data, is similar to ours in performing style transfer via monolingual translation and being applied to formality transfer. In all following evaluations, the model compared to is NMT Combined, which showed the best overall scores in Rao and Tetreault's experiments. 

Following existing work, we employ a CNN classifier, using Lee's implementation\footnote{\url{https://github.com/DongjunLee/text-cnn-tensorflow}} of the architecture proposed by \citet{kim_cnn}. As training data, we use the train split of the GYAFC corpus. 
(See details on model hyperparameters in Appendix \ref{apx3}.) However, the classifier only achieves validation accuracy of 0.79. 

For percentages of test split sentences translated by the two models which were classified as belonging to the target styles, see Table~\ref{tab-classifier-both-dirs}.\footnote{It should be noted that the primary direction in experiments of \citep{sup5} was informal to formal.}

Our model achieves lower scores than the baseline, but shows noticeable movement in the desired direction: 
after applying our model, around 15\% more sentences compared to the original test sets were classified as the target style (in both style transfer directions). 
The classifier was trained on the GYAFC dataset, which is out of domain for our model, but our system still manages to show a noticeable improvement in style transfer scores when compared to source texts.

\begin{table}[]
\centering
\captionof{table}{Percentages of sentences classified as the target style in original GYAFC test sets of the opposite style, and test sets translated into the target style by Rao and Tetreault's (RT) model and by zero-shot machine translation (ZSMT) model. Higher scores for ZSMT/RT models indicate better performance in terms of style transfer.}
\begin{tabular}{c|ccc}
\toprule
     & Original & ZSMT & RT\\
\midrule
fml $\rightarrow$ inf & 8.76 & 23.32  & \textbf{43.60} \\
inf $\rightarrow$ fml & 23.11 & 38.68 & \textbf{77.00}\\
\bottomrule
\end{tabular}
     \label{tab-classifier-both-dirs}
\end{table}


\subsection{Manual Evaluation}
\label{manual-eval}

As automatic evaluation fails to take into account several important aspects of style transfer quality, we perform manual comparison to the model of \citet{sup5} as well.

We evaluate a subset of the test split of the GYAFC corpus, randomly choosing 25 sentences from each domain (Entertainment~\&~Music and Family \& Relationships) and style transfer direction, 100 sentences in total.

The evaluators are presented with the original sentence and outputs of the two systems, in random order, thus, they do not know which output was produced by which system. 3 people participated in the evaluation.\footnote{All annotators are fluent, but non-native speakers of English.} For each output, they were asked: \textit{(i)} to rate how fluent the sentence is, on a scale of 1 to 4, \textit{(ii)} how similar in meaning it is to the source text (1-4), \textit{(iii)} whether the sentence is more formal than the original, more informal, or neither, \textit{(iv)} what differences are there from the original text (with options such as "lexical", "word order", "contractions"; the evaluators are given examples of different types of changes), \textit{(v)} which of the two outputs is better in general: alters style in the right direction, while at the same time preserving meaning of the original text and being fluent (there is an option "no preference", for equally good or equally bad outputs). 

Average fluency and meaning preservation scores of the two systems are in Table \ref{tab-fluency-meaning-comparison}. While both systems score very high on both aspects, zero-shot machine translation is better in both.

\begin{table}[]
\centering
\captionof{table}{Fluency and meaning preservation of Rao and Tetreault's (RT) and zero-shot machine translation (ZSMT) systems, as judged by humans}
\begin{tabular}{c|cc}
\toprule
     & RT & ZSMT \\
\midrule
Fluency (1-4) & 3.67 & \textbf{3.82} \\
Meaning preservation (1-4) & 3.61 & \textbf{3.85} \\
\bottomrule
\end{tabular}
     \label{tab-fluency-meaning-comparison}
\end{table}

In the assessment of style transfer direction, each annotator's vote counted as a +1 if they marked a translated sentence as more formal, and as~-1~if more informal. If the average of votes for a sentence is negative, we consider the sentence more informal than the original text, more formal if it is positive, and if it equals 0, no change in style has occurred. The percentages of sentences in which the direction of style transfer corresponded to the intended one, was wrong, or not present, are given in Table \ref{tab-style-direction-comparison}. Our system tends to be more conservative: it has more sentences where humans detected no changes in style at all. Rao and Tetreault's system shows style transfer in the intended direction more often.

\begin{table}[]
\centering
\captionof{table}{Proportions of sentences where direction of style transfer was found by human annotators to be correct, wrong, and not present, in outputs of Rao and Tetreault's (RT) and zero-shot machine translation (ZSMT) systems}
\begin{tabular}{c|cc}
\toprule
     & RT & ZSMT \\
\midrule
Right direction & \textbf{0.71} & 0.58 \\
Wrong direction & \textbf{0.04} & 0.05 \\
No change in style & 0.25 & 0.37 \\
\bottomrule
\end{tabular}
     \label{tab-style-direction-comparison}
\end{table}

Next, the overall preference was assessed. In 49\% of cases, outputs of our system were preferred by the average annotator. The RT system was preferred in 27\% of cases, and in 24\% of cases no preference was expressed.

Finally, we examined the types of changes reported by the annotators. The most frequent type for our system was contractions (e.g. \textit{I have been} vs. \textit{I've been}), reported by at least one annotator in 38 cases. The next most frequent changes are lexical substitutions (e.g. 
\textit{sure} vs. \textit{certainly}) with 33 cases, punctuation (e.g. \textit{no!!!} vs. \textit{no.}) with 22 cases and missing or added words or phrases with 20 cases. Less common are grammatical changes (e.g. replacing \textit{no one's gonna} with \textit{no one will}
), reported in 8 sentences, and changes in word order, reported in 5 sentences. In 27 sentences out of 100 the annotators did not see any difference from the source sentence, compared to 10 sentences in the baseline output. See Appendix~\ref{apx4} for similar statistics on the RT system. 


In conclusion of this section, while being more conservative in terms of transferring style, our system wins in both fluency and meaning preservation, mostly succeeds in transferring style in the desired direction, and, most importantly, is generally preferred by human evaluators to a baseline trained in a supervised manner and using costly manually constructed data, while performing zero-shot style transfer without any parallel examples.

\subsection{Cross-lingual Style Transfer}

Being able to translate between languages and also to modify the output to match the desired style allows the model to essentially perform domain adaptation. When translating from a language which has no formal "you" (English) into one that does (Estonian or Latvian), it will quite consistently use the informal variant when the target style is OpenSubtitles and the formal when the target style is Europarl (\textit{you rock} $\rightarrow$ \textit{sa rokid/te rokite}). The model is also quite consistent in use of contractions in English (\textit{es esmu šeit} $\rightarrow$ \textit{I am here/I'm here}). 
Some lexical substitutions occur: \textit{need on Matti lapsed.} $\rightarrow$ \textit{those are Matt's kids./these are Matt's children.} Word order may change: \textit{Where is Anna's bag?} is \textit{Kus on Anna kott?} in the more formal variant, and \textit{Kus Anna kott on?} in the more informal. This feature is useful, but out of scope of this article, as we focus on monolingual applications.

\section{Grammatical Error Correction}
\label{gec}

Next we move on to evaluating the same model's performance in the GEC task: for example,
for the English input ``huge fan I are'', the expected output would be ``I am a huge fan''. This section's goal is to systematically check, how reliable our model's corrections are for each kind of grammatical error.

Naturally, the model only copes with some kinds of errors and fails on others -- for instance, it can be expected to restore the correct word order, as long as it does not affect the perception of the meaning, since language models (and NMT decoders) are good at ordering a sequence of known tokens correctly. On the other hand, we do not expect orthographic variations like typos to be fixed reliably, since they affect the sub-word segmentation of the input and can thus hinder translation.

Below we present qualitative and quantitative analysis of our model's GEC results, showing its overall performance, as well as which kinds of errors are handled reliably and which are not.\footnote{Although GEC does not require any distinction in text style, the core idea of this article is to also perform style transfer with the same multilingual multi-domain model. That only means that for GEC we have to select an output domain/style when producing error corrections.}

\subsection{Test Data and Metrics}

We use the following error-corrected corpora both for scoring and as basis for manual analysis:
\begin{itemize}
    \item for English: CoNLL-2014 \cite{conll2014} and JFLEG \cite{jfleg} corpora
    \item for Estonian: the Learner Language Corpus \cite{rummo2017error}
    \item for Latvian: the Error-annotated Corpus of Latvian \cite{deksne2014error}
\end{itemize}
All of these are based on language learner (L2) essays and their manual corrections.

To evaluate the model quantitatively we used two metrics: the Max-Match (M$^2$) metric from the CoNLL-2014 shared task scorer, and the GLEU score \cite{gleu} for the other corpora. The main difference is that M$^2$ is based on the annotation of error categories, while the GLEU score compares the automatic correction to a reference without any error categorization.

\subsection{Results}

The M$^2$ scores are computed based on error-annotated corpora. Since error annotations were only available for English, we calculated the scores on English CoNLL corpus, see Table~\ref{tab-m2}).
\begin{table}[h]
\centering
\captionof{table}{M$^2$ scores on the CoNLL corpus, including precision and recall.}
\begin{tabular}{c|ccc}
  & prec. & recall & M$^2$ \\
\hline
Our model & 33.4 & 27.9 & 32.1\\
Felice, 2014\nocite{felice2014grammatical} & 39.7 & 30.1 & 37.3 \\
Rozovskaya, 2016\nocite{rozovskaya2016grammatical} & 60.2 & 25.6 & 47.4 \\
Rozovskaja (cl) & 38.4 & 23.1 & 33.9 \\
Grundkiewicz, 2018 \nocite{grundkiewicz2018near} & 83.2 & 47.0 & 72.0 \\
\end{tabular}
      \label{tab-m2}
\end{table}

Our model gets the M$^2$ score of 32.1. While it does not reach the score of the best CoNLL model \cite{felice2014grammatical} or the state-of-the-art \cite{grundkiewicz2018near}, these use annotated corpora to train. Our results count as restricted in CoNLL definitions and are more directly comparable to the classifier-based approach trained on unannotated corpora by \citet{rozovskaya2016grammatical}, while requiring even less effort.

\begin{table*}[t]
\centering
\captionof{table}{GLEU scores for all three languages. No scores have been previously reported elsewhere for Estonian and Latvian. ``Original'' scores correspond to non-processed text without any corrections; ``informal'' and ``formal'' results are based on OpenSubtitles/Europarl domains, correspondingly.}
\begin{tabular}{c|cccc}
  & original & informal model & formal model & best known\\
\hline
English (JFLEG) & 40.5 & 44.1 & \textbf{45.9} & 61.5 \\
Estonian & 27.0 & \textbf{38.1} & 37.8 & -\\
Latvian L2 & 59.7 & 44.7 & \textbf{45.1} & -\\
\end{tabular}
      \label{tab-gleu}
\end{table*}

The GLEU scores can be seen in Table~\ref{tab-gleu}. We calculated GLEU for both formal (Europarl) and informal (OpenSubtitles) style models for all three languages. For English our model's best score was 45.9 and for Estonian it was 38.1. Latvian corrected output in fact got worse scores than the original uncorrected corpus, which can be explained by smaller training corpora and worse MT quality for Latvian (see Table~\ref{tab-bleu}). 

\subsection{Qualitative Analysis}

We looked at the automatic corrections for 100 erroneous sentences of English and Estonian each as well as 80 sentences of Latvian. The overall aim was to find the ratio of sentences where
(1) all errors have been corrected
(2) only some are corrected
(3) only some are corrected and part of the meaning is changed and
(4) all meaning is lost.

Analysis was done separately for four error types: spelling and grammatical errors, word choice and word order. In case a sentence included more that one error type it was counted once for each error type. For English the first two types were annotated in the corpus, the rest were annotated by us, separating the original third error category into two new ones; results shown in Table~\ref{tab-gec-categories}.

\begin{table}[h!]
\centering
\captionof{table}{GEC results by error types; ``grammar'' stands for grammatical mistakes, ``lex'' stands for lexical choice and ``order'' -- for word order errors.}
\begin{tabular}{c|cccc|cccc}
& \multicolumn{4}{c|}{Estonian} & \multicolumn{4}{c}{English}\\
& 1 & 2 & 3 & 4 & 1 & 2 & 3 & 4\\
\hline
spelling & 12 & 5 & 2 & 0 & 12 & 7 & 4 & 2\\
lex & 35 & 12 & 18 & 12 &31 & 5 & 5 & 2\\
grammar & 28 & 8 & 8 & 0 & 23 & 13 & 3 & 0\\ 
order & 26 & 5 & 2 & 0 & 2 & 1 & 0 & 0\\
\hline
overall &  29 & 32 & 27 & 12 & 19& 42& 7& 2\\
\end{tabular}
      \label{tab-gec-categories}
\end{table}

Not all English sentences included errors. 30 sentences remained unchanged, out of which 17 had no mistakes in them. For the changed sentences 87\% were fully or partially corrected. In case of Estonian, where all sentences had mistakes, 61 out of the 100 sentences were fully or partially corrected without loss of information. 12 sentences became nonsense, all of which originally had some lexical mistakes. For English the results are similar: the most confusing type of errors that leads to complete loss of meaning is word choice. On the other hand, this was also the most common error type for both languages and errors of that type were fully corrected in 45\% of cases for Estonian and 72\% for English. Using words in the wrong order is a typical learner's error for Estonian that has rather free word order. It is also difficult to describe or set rules for this type of error. Our model manages this error type rather well, correcting 79\% of sentences acceptably, only losing some meaning in 2 such cases.

A similar experiment using 80 Latvian sentences yielded 17 fully corrected sentences, 15, 22 and 26 respectively for the other categories. As the Latvian model is weaker in general, this also leads to more chances of losing some of the meaning; we exclude it from the more detailed analysis and focus on English and Estonian.

Our model handles punctuation, word order mistakes and grammatical errors well. In the following example the subject-verb disagreement in English (\ref{ex:agr-eng}) and verb-object disagreement in Estonian (\ref{ex:agr-est}) have been corrected:  
\begin{enumerate}
  \item 
  \begin{enumerate} \label{ex:agr-eng}
    \item[err:] {“When price of gas goes up , the consumer \textbf{do not} want buy gas for fuels”} 
    \item[fix:]{“When the price of gas goes up, the consumer \textbf{doesn't} want to buy gas for fuels”}
  \end{enumerate}
  \item 
  \begin{enumerate} \label{ex:agr-est}
  \item[err:]{“Sellepärast ütleb ta filmi lõpus, et tahab oma \textbf{unistuse} tagasi”}
  \item[fix:]{“Sellepärast ütleb ta filmi lõpus, et tahab oma \textbf{unistust} tagasi”}
  \item[gloss:]that's-why says he film at-end, that (he)-wants his-own dream$_{(genitive/partitive case)}$
  \end{enumerate}
\end{enumerate}

Sentences that include several error types are generally noticeably more difficult to correct. Depending on the error types that have been combined our model manages quite well and corrects all or several errors present. Example~\ref{ex:complex-est} includes mistakes with word order and word choice: the argument \textit{"vabaainetele"} (\textit{to elective courses}) here should precede the verb and the verb \textit{"registreeruma"} (\textit{register oneself}) takes no such argument. Our model corrects both mistakes while also replacing the\textit{"seejärel"} (\textit{after that}) with a synonym.

\begin{enumerate}
  \setcounter{enumi}{2}
  \item 
  \begin{enumerate} \label{ex:complex-est}
  \item[err:] {\textbf{Seejärel} pidi igaüks \textbf{ennast registreeruma} vabaainetele.
  \item[gloss:]then had-to everyone oneself register-oneself to-free-courses}
  \item[fix:] {\textbf{Siis} pidi igaüks \textbf{end} vabaainetele \textbf{registreerima}.
    \item[gloss:]then had-to everyone oneself to-free-courses register}
  \end{enumerate}
\end{enumerate}

Results also show two main downsides of our model. First, it is much less reliable with lexical mistakes and typos: 
for example, in some sentences a misspelled word is changed into an incorrect form that has a common ending, like \textit{"misundrestood"} to \textit{"misundrested"}. 
Another drawback is that sometimes more changes are applied to the input than strictly necessary: sometimes this meant replacing words with their synonyms (like replacing \textit{``frequently''} with \textit{``often''}), in other cases the changes distorted the meaning (in one case, the Estonian \textit{``öelda'' (say)} was replaced with \textit{``avaldada'' (publish)}, and in another, \textit{``siblings''} was replaced with \textit{``friends''}).




To conclude this section, our model reliably corrects grammatical, spelling and word order errors, with more mixed performance on lexical choice errors and some unnecessary changes of the input, which are sometimes paraphrases, and sometime more loose re-interpretations. The error types that the model manages well can be traced back to having a strong monolingual language model, a common trait of a good NMT model. As the model operates on the level of word parts and its vocabulary is limited, this leads to combining wrong word parts,
sometimes across languages. 

\section{Related Work}
\label{relwork}

\textbf{Style transfer:} 
Several approaches use directly annotated data: for example, \citet{sup1} and \citet{sup2} train MT systems on the corpus of modern English Shakespeare to original Shakespeare. 
\citet{sup5} collect a dataset of 110K informal/formal sentence pairs and train rule-based, phrase-based, and neural MT systems using this data. 


One line of work aims at learning a style-independent latent representation of content while building decoders that can generate sentences in the style of choice \cite{g1,g2,g3,g4,g5,g6, vae, gan}. Unsupervised MT has also been adapted for the task 
\cite{umt1, umt2}. 
Our system also does not require parallel data between styles, but leverages the stability of the off-the-shelf supervised NMT to avoid the hassle of training unsupervised NMT systems and making GANs converge.

Another problem with many current (both supervised and unsupervised) style transfer methods is that they are bounded to solve a binary task, where only two styles are included (whether because of data or restrictions of the approach). 
Our method, on the other hand, can be extended to as many styles as needed as long as there are parallel MT corpora in these styles available.
Notably, \citet{politeness} use side constrains in order to translate in polite/impolite German, while we rely on multilingual encoder representations and use the system monolingually at inference time.

\citet{gbt} translate a sentence into another language, hoping that it will lose some style indicators, and then translate it back into the original language with a desired style tag attached to the encoder latent space. We also use the MT encoder to obtain rich sentence representations, but learn them directly as a part of a single multilingual translation system.

Finally, \citet{related1} also employ machine translation to approach the task of monolingual and cross-lingual style transfer. However, they use supervised data (formal/informal sentence pairs) in training while our method does not. 

\textbf{Grammatical error correction:} 
there have been four shared tasks for GEC with prepared error-tagged datasets for L2 learners of English in the last decade: HOO \cite{dale2011helping, dale2012hoo} and CoNLL \cite{ng2013conll, conll2014}.
Approaches to GEC split into either rule-based approaches, machine learning on error-tagged corpora, MT models on parallel data of erroneous and corrected sentences, or a combination of these \cite{conll2014}. The top model of the CONLL shared task in 2014 used a combined model of rule-based approach and MT \cite{felice2014grammatical}. A more recent development in the same direction is the combination of an error-correcting FST with a neural language model that computes the FST's weights \cite{ngec}. All of these require annotated data or considerable effort to create, whereas our model is much more resource-independent.  

Another focus of the newer research is on creating GEC models without human-annotated resources. For example \citet{rozovskaya2016grammatical} combine statistical MT with unsupervised classification using unannotated parallel data for MT and unannotated native data for the classification model. In this case parallel data of erroneous and corrected sentences is still necessary for MT; the classifier uses native data, but still needs definitions of possible error types to classify -- this work needs to be done by a human and is difficult for some less clear error types.

There has been little work on Estonian and Latvian GEC, all limited to rule-based approaches \cite{liin2009komavigade,deksne2016new}.
For both languages, as well as any low-resourced languages, our model gives a way to do grammatical error correction without the need for error-corrected corpora. 

\section{Conclusions}
\label{conclusions}

We presented a simple approach where a single multilingual NMT model is adapted to monolingual transfer and performs grammatical error correction and style transfer. We experimented with three languages and presented extensive evaluation of the model on both tasks. We used publicly available software and data and believe that our work can be easily reproduced.

We showed that for grammatical error correction our approach reliably corrects spelling, word order and grammatical errors, while being less reliable on lexical choice errors. Applied to style transfer our model is good at meaning preservation and output fluency, while reliably transferring style for English contractions, lexical choice and grammatical constructions. From the practical point of view when applied to both tasks, our approach can be best used by someone who has some command of a language and can thus re-evaluate the model's suggested corrections or style adaptations, but can benefit from suggestions of corrections that they might not have come up with. The main benefit is that no annotated data is used to train the model, thus making it very easy to train it for other (especially under-resourced) languages.


Future work includes exploring adaptations of this approach to both tasks separately, while keeping the low cost of creating such models, exploring character-level models for lexical error corrections and estimating the influence of the number and choice of languages on the performance.


\bibliographystyle{acl_natbib}
\bibliography{acl2019}

\newpage
~
\newpage

\appendix



\section{Model Training: Technical Details}
\label{apx1}
After rudimentary cleaning (removing pairs where at least one sentence is longer that 100 tokens, at least one sentence is an empty string or does not contain any alphabetic characters, and pairs with length ratio over 9) and duplication to accommodate both translation directions in each language pair, the total size of the training corpus is 22.9M sentence pairs; training set sizes per language and corpus are given in Table~\ref{tab-trainsizes}.
Validation set consists of 12K sentence pairs, 500 for each combination of translation direction and corpus. We also keep a test set of 24K sentence pairs, 1000 for each translation direction and corpus.


\begin{table}[h]
\centering
\captionof{table}{Training set sizes (number of sentence pairs)}
\begin{tabular}{c|ccc}
\toprule
      & EN $\leftrightarrow$ ET & EN $\leftrightarrow$ LV    & ET $\leftrightarrow$ LV\\
\midrule
EP & 0.64M & 0.63M & 0.63M \\
JRC & 0.68M & 0.69M & 1.5M \\
EMEA & 0.91M & 0.91M & 0.92M \\ 
OS & 3M & 0.52M & 0.41M \\
\bottomrule
\end{tabular}
      \label{tab-trainsizes}
\end{table}

The data preprocessing pipeline consists of tokenization with Moses tokenizer \cite{moses}, true-casing, and segmentation with SentencePiece \cite{sentencepiece} with a joint vocabulary of size 32 000.

We trained a Transformer NMT model using the Sockeye framework \cite{sockeye}, mostly following the so-called \emph{Transformer base model}: we used 6 layers, 512 positions, 8 attention heads and ReLU  activations  for  both  the encoder  and  decoder; Adam optimizer was used. Source and target token embeddings were both of size 512, and factors determining target language and style had embeddings of size 4. Batch size was set to 2048 words, initial learning rate to 0.0002, reducing by a factor of 0.7 every time the validation perplexity had not improved for 8 checkpoints, which happened every 4000 updates. The  model  converged  during  the  17th epoch,  when  validation  perplexity  has  not improved for 32 consecutive checkpoints. The parameters of a single best checkpoint were used for all translations, with beam size set to 5.

\section{Classifier Training: Technical details}
\label{apx3}

The total number of sentences in the GYAFC train split is 209,000, of which 40,000 are used for validation. We train a convolutional neural network with filter sizes 3, 4, and 5, with 256 filters per size, with ReLU activation and max-pooling. No pre-trained word embeddings are used, instead, embeddings of dimension 300 are learned during training. Adam optimizer is used, with initial learning rate $5\times10^{-4}$. The convolutional layer is followed by one feed-forward layer. Dropout rate is set to 0.5. The classifier achieves validation accuracy of 0.79.

\section{Types of Changes Made}
\label{apx4}

Table~\ref{tab-types-comparison} shows how often different types of changes to the source text were reported in the output of our zero-shot system and the supervised system of \citet{sup5}.

\begin{table}[]
\centering
\captionof{table}{Proportion of output sentences of the two systems in which varying types of changes were reported by human annotators (RT -- Rao and Tetreault's benchmark, ZSMT -- zero-shot machine translation system)}
\begin{tabular}{p{4cm}|cc}
\toprule
     & ZSMT & RT \\
\midrule
Contractions & 0.38 & 0.36 \\
Lexical substitutions & 0.33 & 0.46 \\
Punctuation & 0.22 & 0.49 \\
Missing or added words or phrases & 0.20 & 0.39 \\
Grammatical variations & 0.08 & 0.16 \\
Word order & 0.05 & 0.07 \\
No difference & 0.27 & 0.10 \\
\bottomrule
\end{tabular}
     \label{tab-types-comparison}
\end{table}

\section{Output Examples}
\label{apx2}

We present more examples of translation of code-switched input segments, error correction and style transfer in English, Estonian and Latvian, informal (inf) and formal (fml) output style:

\newcommand{\arrr}{$\rightarrow$}
\begin{table}[h]
    \centering
    \begin{tabular}{p{13mm}|p{28mm}|p{30mm}}
        ET\arrr{}ET (fml) & Mida sa teed? & Mida te teete? \\
        \midrule
        ET\arrr{}ET (fml) & Milleks tulid? & Miks te tulite? \\
        \midrule
        EN\arrr{}EN (inf) & I will reimburse you. & I'll pay you back. \\
        \midrule
        EN\arrr{}EN (inf) & That is correct. & That's right. \\
        \midrule
        EN\arrr{}EN (fml) & It's a pretty important part of the deal. & It is a fairly important part of the deal. \\
        \midrule
        EN\arrr{EN} (fml) & He big boss, he make much money. & He big boss, he makes a lot of money. \\
        \midrule
        ET\arrr{}ET (fml) & Ta olen suured poisi. & Ta on suur poiss.\\
        \midrule
        LV\arrr{}LV (fml) & Mums esi grūts. & Mums ir grūti. \\
        \midrule
        EN\arrr{}ET (inf) & You are bad & Sa oled paha! \\
        \midrule
        EN\arrr{}ET (fml) & You are bad & Te olete pahad! \\
        \midrule
        ET\arrr{}LV (inf) & Sinu plaan on jama! & Tavs plāns ir stulbs! \\
        \midrule
        ET\arrr{}LV (fml) & Sinu plaan on jama! & Jūsu plāns ir nejēdzīgs! \\
        \midrule
        ?\arrr{}EN (inf) & Kes you esi? & Who are you? \\
        \midrule
        ?\arrr{}LV (inf) & Kes you esi? & Kas tu esi? \\
        \midrule
        ?\arrr{}ET (inf) & Kes you esi? & Kes sa oled? \\
    \end{tabular}
\end{table}




\end{document}